\documentclass[conference]{IEEEtran} 
\IEEEoverridecommandlockouts
\usepackage[utf8]{inputenc}
\usepackage[T1]{fontenc}

\usepackage{orcidlink}
\usepackage{graphicx}
\usepackage{url}
\usepackage{hyperref}
\usepackage{booktabs}
\usepackage{multirow}
\usepackage{longtable}
\usepackage{diagbox}
\usepackage{algorithm}
\usepackage{algpseudocode}

\graphicspath{{images/}}

\title{State Compression in Two-Agent LLM Relays: A Closed-World Study of Constraint Preservation}

\author{ 
    \IEEEauthorblockN{Anantha Sharma\orcidlink{0000-0002-9064-3362}}
    \IEEEauthorblockA{\textit{The A-Team} \\
    \textit{Synechron Inc.}\\
    Charlotte, USA}
\and
    \IEEEauthorblockN{Sheeba Elizabeth John \orcidlink{0009-0003-8384-7521}}
    \IEEEauthorblockA{\textit{The A-Team} \\
    \textit{Synechron Inc.}\\
    New York, USA}
\and
    \IEEEauthorblockN{Kaarthik Senthil Kumar \orcidlink{0009-0007-1461-1231}}
    \IEEEauthorblockA{\textit{The A-Team} \\
    \textit{Synechron Inc.}\\
    Charlotte, USA}
\and
    \IEEEauthorblockN{Saratsuhas Vijayababu \orcidlink{0009-0006-6682-7918}}
    \IEEEauthorblockA{\textit{The A-Team}\\
    \textit{Synechron Inc.}\\
    New York, USA}
\thanks{All authors contributed equally to this work.}
}

\begin{document}

\maketitle

\begin{abstract}
Long-running Large Language Model (LLM)-based agents often accumulate large intermediate traces containing audits, eliminations, and numeric calculations. In practice, this state is compressed before handing it to a downstream decision step, creating an information bottleneck in which small omissions can break strict numeric or categorical constraints. This paper evaluates hand-off compression in a closed-world travel-planning relay with two LLM agents. A Researcher audits a fixed inventory of hotels and flights for 50 goal instances, and a Booker selects a hotel--flight pair using only the goal and the hand-off payload, with the inventory withheld. We compare four hand-off conditions: no compression, narrative summarization, schema-constrained JSON extraction, and embedding-based pruning. Exhaustive enumeration over the fixed inventory provides exact feasible and optimal labels. Results show that hand-off representation strongly affects downstream feasibility under a small decision model. JSON extraction achieves the highest feasibility accuracy at 0.96, while narrative summarization, despite producing the smallest compressed hand-off payload, degrades feasibility to 0.48. Embedding-based pruning matches the uncompressed control on feasibility at 0.88 without an additional generative compression call. These findings indicate that constraint checking benefits from structured and auditable hand-off representations rather than relying on brevity alone.
\end{abstract}

\section{Introduction}
A common pattern with agentic systems is to solve multi-step tasks by generating long intermediate working memory traces \cite{wu2026memory, zhang2026lightweight} that include audits, eliminations, calculations, and rationale. As these traces grow, agentic systems face practical constraints from limited context windows of the underlying models and rising inference cost. A common engineering response compresses \cite{kang2025acon} the agent state before handing it to a downstream decision component \cite{li2023selectivecontext,jiang2023llmlingua,jiang2024longllmlingua,kang2025acon}. This hand-off creates an information bottleneck, since compression must remove verbosity while preserving decision-critical facts, particularly tight numerical and categorical constraints required for valid decision-making. Small distortions can be operationally significant. An approximate budget or a missing amenity can shift a downstream choice from valid to invalid or lead to reconciliation gaps.

This paper studies state compression at a single, controlled hand-off bottleneck using a \textbf{closed-world travel planning} benchmark. We define state compression as transforming an upstream trace into a smaller representation intended to preserve decision-critical constraints. We evaluate a two-stage relay protocol. A Researcher agent produces an exhaustive per-option audit over a fixed inventory, explicitly checking each hotel and each flight against the goal’s hard constraints and recording pass/fail evidence, yielding a long trace \cite{liu2024lostinthemiddle}. A Booker agent receives only the original goal and a compressed form of that trace and selects a hotel--flight pair intended to satisfy the constraints.

We compare four hand-off conditions, an uncompressed control and three compression strategies: narrative summarization \cite{kang2025acon}, schema-constrained JSON extraction, and embedding-based pruning that filters sentences using cosine similarity in embedding space. Ground truth is computed by exhaustive enumeration of all hotel--flight pairs over the fixed inventory, which enables objective scoring even when multiple feasible solutions exist. Feasibility accuracy is treated as the primary outcome, optimal match accuracy and optimality gap are secondary task metrics, and compression ratio and tokens per run are reported as efficiency measures. Embedding-space similarity is used as a representation diagnostic. Results quantify trade-offs between compression aggressiveness and downstream reliability, and identify failure patterns in which compression retains candidate mentions while dropping elimination evidence, which can lead to invalid selections.

\section{Methodology}
This work evaluates compression of agent state at a controlled hand-off between two LLM agents. A closed-world travel planning benchmark was used so that feasible solutions could be computed exactly. Each run followed a fixed relay sequence: an upstream Researcher agent produced an exhaustive audit trace over a constant inventory of flights and hotels, a compression layer transformed that trace using one of several techniques, and a downstream Booker agent made the final booking decision using only the goal and the compressed hand-off.

\subsection{Closed-World Dataset and Goal Specification}
A closed-world travel planning dataset was constructed for London consisting of a fixed inventory of 10 hotels and 10 flights. Each hotel record included an identifier, price per night in USD, distance to the city center in kilometers, star rating, and a finite set of amenities represented as canonical tokens. Each flight record included an identifier, airline, departure time window, cabin class, and price in USD. Additional flight attributes such as layovers and refundability were included as distractor fields and were not enforced as hard constraints. All prices were defined for a single traveler and a one-night stay, and total trip cost was computed as the sum of one flight and one hotel night.

The evaluation set contained 50 goal instances. Each goal specified a traveler persona and a set of hard constraints used for scoring: a total budget cap, a must-have hotel amenity, a maximum hotel distance to the city center, a preferred departure time window, and a required cabin class. Goals were designed to span a range of constraint tightness, including near-miss cases where feasibility depended on small budget margins. The fixed inventory remained constant across all runs, while goals varied. This ensured that observed differences primarily reflected the hand-off representation and compression method rather than changes in the underlying world state.

\subsection{Ground Truth Generation}
Ground truth was generated deterministically by exhaustive enumeration over the closed-world inventory. For each goal instance, we considered hotel--flight pairs formed from the inventory and filtered them by the hard constraints defined in the goal, including total budget, required hotel amenity, maximum hotel distance, preferred departure window, and required cabin class. Amenities were matched by exact token membership. A pair was feasible if the hotel and flight satisfied the categorical constraints and the combined cost of one hotel night plus one flight did not exceed the budget. Let $g$ denote a goal instance and let $F_g$ denote the set of all feasible hotel--flight pairs for $g$ under the hard constraints. If $F_g = \emptyset$, the goal was labeled unsatisfiable and the expected output was \texttt{NONE} for both identifiers.

A single best pair was defined to support secondary optimality analysis in addition to the feasible set. Let $t_g^\ast \in F_g$ denote the top-ranked feasible pair under a deterministic ordering. Feasible pairs were ranked by minimum total trip cost, then minimum hotel distance to the city center, then maximum hotel star rating, with a final tie-break by lexicographic ordering of (\texttt{hotel\_id}, \texttt{flight\_id}). The selected top-ranked pair $t_g^\ast$ was recorded as the optimal pair for goal $g$, and the full feasible set $F_g$ was recorded to support evaluation when multiple feasible solutions existed. This ground-truth generation procedure was implemented as a standalone script and was independent of the LLM agents, which ensured reproducible labels across runs.

\subsection{Two-Agent Relay Protocol}
\begin{figure}[t]
\centering
\includegraphics[width=\linewidth]{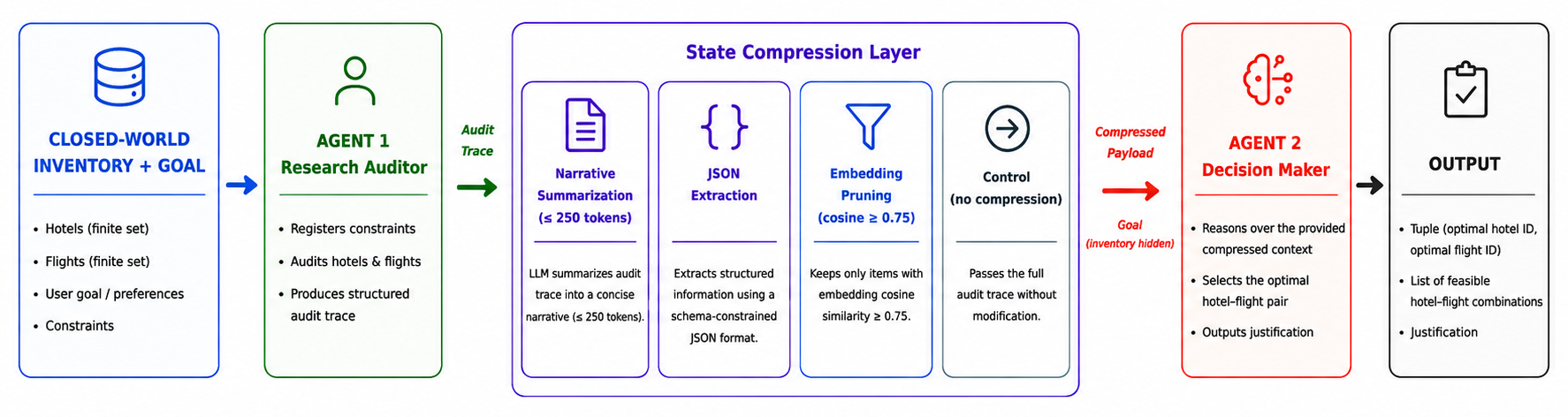}
\caption{Evaluating hand-off compression methods in a closed-world two-agent relay (Researcher Agent $\rightarrow$ compression $\rightarrow$ Booker Agent).}
\label{fig:architecture}
\end{figure}

Each experimental run followed a fixed two-agent relay designed to isolate the hand-off bottleneck. The upstream Researcher agent received the goal instance together with the full closed-world inventory of flights and hotels. Its task was to produce a long, exhaustive audit trace that evaluated every hotel and every flight against the hard constraints used for scoring, namely total budget, must-have amenity, maximum distance to the city center, flight departure window, and cabin class. The Researcher agent was instructed to operate under a strict closed-world rule, using only the provided inventory and avoiding any external assumptions. Both agents used \emph{GPT-4.1-nano} \cite{azure_openai_gpt41_nano}, a small model configuration, and embeddings were computed using \emph{text-embedding-3-small} \cite{azure_openai_text_embedding_3_small}. Fig.~\ref{fig:architecture} summarizes the closed-world two-agent relay and the three compression variants evaluated at the hand-off bottleneck.

The Researcher trace was then passed through a compression layer that produced a transformed hand-off payload under one of the evaluated conditions. The downstream Booker agent received the original goal and only the compressed payload. The closed-world inventory was not provided to the Booker agent, so the booking decision depended entirely on what survived the compression step. The Booker agent was required to enumerate candidate hotel--flight combinations explicitly supported by evidence in the payload and to select a single recommendation using the same deterministic ranking used by the ground-truth generator, prioritizing minimum total trip cost, then minimum hotel distance, then maximum hotel star rating, with stable identifier ordering used only as a final tie-break. The Booker agent was not permitted to introduce hotel or flight identifiers that did not appear in the upstream payload. This restriction follows the broader goal of reducing ungrounded generations (hallucinations) in downstream components \cite{zerog2024}.

A no-compression control condition was included, where the full Researcher trace was passed to the Booker agent unchanged. All runs used deterministic decoding with temperature set to 0 and a fixed seed per run. Prompts and response schemas were held constant across conditions. Each run was logged end-to-end, including the Researcher output, the compressed payload, and the Booker decision, supporting reproducible analysis of compression-induced information loss and downstream constraint violations.

\subsection{Compression Conditions}
We evaluated four hand-off conditions. Prompt and context compression have been explored in prior work using redundancy pruning and learned compressors \cite{li2023selectivecontext,jiang2023llmlingua,jiang2024longllmlingua,kang2025acon}, which motivates our pruning approach. The control condition performed no compression and passed the full Researcher trace to the Booker agent unchanged. Three compression methods were then applied to the same Researcher trace. The first method produced a narrative text summary capped at 250 words. The second method performed schema-constrained structured extraction, which transformed the trace into a JSON object that recorded the constraint register, per-option evaluations and elimination evidence in a normalized format. Prior work on constrained decoding for structured outputs motivated enforcing output validity when generating structured targets \cite{scholak2021picard}. The third method applied embedding-based pruning, splitting the Researcher trace into sentences and retaining sentences based on cosine similarity in embedding space to the goal and to an elimination-oriented query, while explicitly preserving identifier and verdict sentences to reduce accidental loss of decision-critical evidence. A small threshold sweep was performed during development, and a cosine threshold of 0.75 was selected for the main study as a stable operating point.

Compression was implemented as a deterministic middleware layer rather than an additional agent in the relay. This avoided introducing an extra decision-maker whose stochastic behavior could confound attribution of errors. Each run and condition recorded the compressed payload, token usage (as a proxy for inference cost), and embedding computations, and the resulting payload became the sole evidence available to the downstream Booker agent.

\subsection{Evaluation Metrics and Logging}
Evaluation combined task-level correctness with efficiency and representation-level measurements recorded at the hand-off. Feasibility accuracy served as the primary outcome. Let $t_g$ denote the Booker agent selection for goal $g$. A run was scored as correct if $t_g \in F_g$. When $F_g = \emptyset$, the output \texttt{NONE} for both identifiers was treated as correct. Optimal match accuracy served as a secondary metric and was defined as whether $t_g = t_g^\ast$. Alongside this, an optimality gap was computed \emph{conditional on a feasible selection} (i.e., $t_g \in F_g$) as $\Delta_g = C(t_g) - C(t_g^\ast)$, where $C(\cdot)$ denotes total trip cost in USD computed from the closed-world inventory. $\Delta_g$ is undefined otherwise.

Compression ratio was defined as the number of tokens in the hand-off payload divided by the number of tokens in the original Researcher trace. Token usage was recorded per run, accounting for the Researcher and Booker calls and any additional calls required by the compression method. Representation drift was measured using cosine similarity in embedding space. Embeddings were computed using \texttt{text-embedding-3-small} for the goal text, the Researcher trace, and each compressed payload, and similarities were recorded between the Researcher trace and the compressed payload and between the compressed payload and the goal. These similarity measures were treated as proxies for representation shift and do not guarantee preservation of numeric thresholds or categorical constraints. 

All runs were logged with the complete Researcher trace, compressed payload, Booker output, token usage, trip-cost quantities used to compute $\Delta_g$, and the above metrics. Logs also included goal identifiers and condition labels. This enabled paired comparisons across compression methods on the same set of goals and supported failure-mode analysis at the level of individual constraints.

\section{Results}
We conducted an empirical comparison of four hand-off conditions on the London closed-world benchmark (N=50 goals). Feasibility accuracy was treated as the primary outcome, with optimal match accuracy and optimality gap as secondary task metrics. Compression ratio and token usage were reported to characterize efficiency, and embedding-based similarity diagnostics were used to characterize representation shift. We report 95\% bootstrap confidence intervals (10{,}000 resamples over goals) and use exact McNemar tests for paired comparisons of feasibility outcomes.

\subsection{Overall Comparison}
Table~\ref{tab:overall} reports overall performance across conditions. Tokens per run is the total token usage per run across the Researcher and Booker calls, including any additional calls required by the compression method.

\begin{table}[b]
\caption{Overall performance across four hand-off conditions ($N{=}50$ goals): 95\% bootstrap CI.}
\begin{center}
\begin{tabular}{|p{1.6cm}|p{1.5cm}|p{1.5cm}|p{1.4cm}|p{0.9cm}|}
\hline
\textbf{Method} & \textbf{Feasible \ Accuracy} & \textbf{Optimal\ Accuracy} & \textbf{Compression \ Ratio} & \textbf{Tokens per run} \\
\hline
Control (no compression) & 0.88 [0.78,0.96] & 0.70 [0.58,0.82] & 1.000 & 6{,}727 \\
\hline
Narrative Summary & 0.48 [0.34,0.62] & 0.42 [0.28,0.56] & 0.333 & 7{,}569 \\
\hline
JSON Extraction & 0.96 [0.90,1.00] & 0.74 [0.62,0.86] & 0.738 & 9{,}651 \\
\hline
Embedding Pruning & 0.88 [0.78,0.96] & 0.68 [0.56,0.80] & 0.826 & 7{,}903 \\
\hline
\end{tabular}
\label{tab:overall}
\end{center}
\end{table}

\begin{figure}[t]
\centerline{\includegraphics[width=1\linewidth]{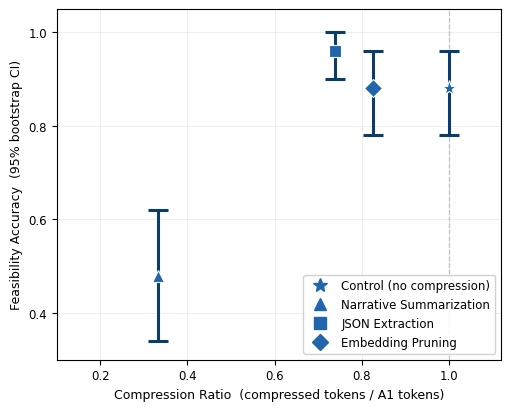}}
\caption{Feasibility accuracy vs.\ compression ratio for the hand-off representations setting: 95\% bootstrap CI over 50 goals. Lower compression ratio indicates a smaller hand-off payload relative to the Researcher trace.}
\label{fig:tradeoff}
\end{figure}

JSON extraction achieved the highest feasibility accuracy (0.96), followed by the control and embedding pruning (0.88 each), while the 250-word narrative summary substantially degraded feasibility (0.48). Fig.~\ref{fig:tradeoff} visualizes this feasibility-compression trade-off and highlights that JSON attains the highest feasibility while reducing the hand-off footprint relative to the uncompressed control. The narrative summary condition produced the most aggressive compression (ratio 0.333) but also the largest feasibility loss, consistent with a bottleneck in which constraint-critical evidence is omitted. By contrast, JSON extraction retained a larger payload (ratio 0.738) and improved feasibility relative to the uncompressed trace. This indicates that a normalized, schema-constrained hand-off can reduce downstream error even when it is less compact.

On optimal match accuracy, JSON extraction again ranked highest (0.74), followed by the control (0.70) and embedding pruning (0.68), with narrative summary lowest (0.42). 

Table~\ref{tab:gap} reports the optimality gap $\Delta_g=C(t_g)-C(t_g^\ast)$ in USD, computed only when the goal is satisfiable and the Booker output is feasible ($t_g \in F_g$), since $\Delta_g$ is undefined otherwise.

Across conditions, mean gaps remain in the tens of dollars, which indicates that many non-optimal outputs are still feasible but reflect ranking errors among feasible alternatives rather than constraint violations. Among feasible outputs, JSON extraction attains the lowest mean gap at \$15.12 [\$2.20,\ \$37.44] and the largest feasible sample size (highest $N{=}41$), consistent with more reliable selection once feasibility is satisfied. Narrative summarization has the smallest $N$ ($N{=}19$) and this reflects frequent feasibility failures that prevent gap measurement.

These results separate \emph{constraint failure} from \emph{ranking error}. Once feasibility is achieved, the remaining mistakes are typically sub-optimal choices among feasible alternatives rather than invalid bookings. This complements the feasibility-compression trade-off in Fig.~\ref{fig:tradeoff}, where structured JSON achieves the highest feasibility under context reduction.

\subsection{Failure Modes}
Table~\ref{tab:failures} breaks down errors by type. Narrative summarization produced the most infeasible outcomes (26/50), including frequent false-\texttt{NONE} decisions (8/50), where the Booker declared no valid plan despite the existence of feasible pairs in the closed-world inventory. JSON extraction eliminated false-\texttt{NONE} in this run (0/50) and reduced feasibility failures to 2/50. This indicates that preserving explicit elimination and constraint evidence in a structured form is critical for downstream feasibility. Control and embedding pruning exhibited similar feasibility failure counts (6/50), but differed in token usage and compression footprint (Table~\ref{tab:overall}).

Exact McNemar tests on paired per-goal feasibility outcomes (\texttt{feasible\_success}) supported these differences. Narrative summarization was significantly worse than control, JSON extraction, and embedding pruning ($p<0.001$ with Bonferroni correction), while JSON extraction did not significantly differ from control in this run ($p=0.2891$).

\begin{table}[t]
\caption{Optimality gap $\Delta_g$ (USD) for feasible outputs only ($t_g \in F_g$): 95\% bootstrap CI.}
\label{tab:gap}
\begin{center}
\footnotesize
\setlength{\tabcolsep}{3pt}
\renewcommand{\arraystretch}{1.15}
\begin{tabular}{|p{1.6cm}|p{2.2cm}|p{0.9cm}|}
\hline
\textbf{Method} & \textbf{Mean Gap (95\% CI)} & \textbf{N} \\
\hline
Control & 17.44 [6.03, 31.54] & 39 \\
\hline
Narrative Summary & 23.42 [0.00, 53.42] & 19 \\
\hline
JSON Extraction & 15.12 [2.20, 37.44] & 41 \\
\hline
Embedding Pruning & 23.92 [9.05, 41.89] & 37 \\
\hline
\end{tabular}
\end{center}
\end{table}

\begin{table}[b]
\caption{Failure breakdown by condition (counts over 50 goals).}
\label{tab:failures}
\centering
\footnotesize
\setlength{\tabcolsep}{3pt}
\renewcommand{\arraystretch}{1.15}
\begin{tabular}{|p{1.6cm}|p{1.0cm}|p{1.0cm}|p{1.0cm}|p{1.2cm}|p{1.2cm}|}
\hline
\textbf{Method} & \textbf{Feas.\ OK} & \textbf{Opt.\ OK} & \textbf{Feas.\ Fail} & \textbf{false-\texttt{NONE}} & \textbf{Wrong-pair} \\
\hline
Control & 44 & 35 & 6 & 3 & 9 \\
\hline
Narrative Summary (250w) & 24 & 21 & 26 & 8 & 3 \\
\hline
JSON  & 48 & 37 & 2 & 0 & 11 \\
\hline
Embedding Pruning 
($\tau{=}0.75$) & 44 & 34 & 6 & 4 & 10 \\
\hline
\end{tabular}
\end{table}

\subsection{Compression Efficiency and Representation Diagnostics}
We also measured embedding-based fidelity (cosine similarity between the Researcher trace and the compressed payload) and goal similarity (cosine similarity between the compressed payload and the goal). Narrative summarization was most goal-aligned (0.676) but lowest in fidelity (0.669), whereas embedding pruning had the highest fidelity (0.898) but the lowest goal similarity (0.449); JSON extraction was intermediate (fidelity 0.745, goal similarity 0.514). These diagnostics do not guarantee constraint preservation, but help characterize representation shift and complement the feasibility--compression trade-off shown in Fig.~\ref{fig:tradeoff}.

\section{Discussion and Analysis}
Observed performance differences reflected how each hand-off representation supported downstream verification under a limited-capacity decision model. In the small-model relay setting, the bottleneck is more pronounced because the Booker agent depends entirely on the hand-off payload and is less able to recover from missing or noisy evidence. Structured extraction achieved the highest feasibility (0.96), narrative summarization produced the lowest feasibility (0.48), and embedding pruning matched the uncompressed control on feasibility (0.88), which suggests that payload length alone did not determine downstream reliability. 
This is consistent with the view that agent reliability often depends on externalized infrastructure and harness design rather than model capability alone \cite{zhou2026externalization}.
Fig.~\ref{fig:tradeoff} highlights this non-monotonic relationship: structured JSON achieves higher feasibility than the control despite a lower compression ratio, while embedding pruning clusters near the control in both feasibility and ratio. From an efficiency standpoint, the gain in feasibility under JSON comes at a higher token budget than the control (Table~\ref{tab:overall}), likely because schema-guided outputs introduce structural overhead (keys and repeated fields) even as they improve verifiability. In contrast, embedding pruning avoids an additional generative compression call and largely preserves original evidence, which enables control-level feasibility with a moderate token footprint.

\subsection{Why Structured Extraction Outperformed the Uncompressed Trace}
JSON extraction produced the highest feasibility accuracy (0.96) even though the payload was not the smallest. This suggests that explicit structure and normalized evidence can matter more than raw context volume under the relay protocol. The uncompressed trace contains extensive narrative content and local calculations, which may increase the chance that a small downstream model misses key disqualifying details or fails to assemble viable combinations \cite{li2024, liu2024lostinthemiddle}. Schema-constrained extraction reduces this burden by presenting constraints and per-option evaluations in a normalized format, which improves verifiability. This interpretation is consistent with the elimination of false-\texttt{NONE} outputs under JSON extraction in this run (0/50), compared to 3/50 under the uncompressed control.

\subsection{Narrative Summarization and Evidence Loss Under Tight Constraints}
Narrative summarization achieved the smallest hand-off payload (compression ratio 0.333), yet it produced the lowest feasibility accuracy (0.48) and the highest rate of false-\texttt{NONE} (8/50). The failure breakdown suggests that the summarizer often removed constraint-critical details needed for strict verification, such as exact prices, distance values, or elimination evidence that distinguishes near-miss options. Goal similarity remained relatively high under summarization, which implies that semantic alignment alone does not guarantee retention of numerical or categorical facts. The feasibility failures therefore appear consistent with a compression regime that preserves intent while discarding details required for closed-world constraint checking.

\subsection{Embedding-Based Pruning as a Token-Efficient Baseline}
Embedding pruning matched the control condition on feasibility accuracy (0.88) while avoiding an additional generative compression call. Fidelity to the Researcher trace remained highest under pruning, which reflects retention of original sentences rather than rephrased summaries. Lower goal similarity under pruning suggests the retained content is not necessarily goal-focused; however, feasibility remained competitive and this shows that literal preservation of evidence can be sufficient when enough disqualifying statements survive the filter. The remaining false-\texttt{NONE} cases under pruning (4/50) highlight sensitivity to thresholding when elimination evidence is sparse.

\subsection{Limitations and Implications}
The study used a closed-world inventory and a fixed task structure, which enabled exact labeling but limited the scope of generalization to open-world retrieval settings. Model capacity also influenced outcomes. Both agents use a small model configuration, which likely amplified differences that could be smaller under stronger downstream models. Practical agent pipelines often mix model sizes, so the observed improvement of structured extraction over the uncompressed trace suggested a useful design implication that smaller decision models may benefit from normalized hand-offs even when raw context is available.

\section{Conclusion}
This paper evaluated state compression at a single hand-off bottleneck in a closed-world two-agent travel-planning relay. A fixed inventory of hotels and flights enabled exact ground-truth labeling of feasible and optimal hotel--flight pairs, which allowed objective comparison of hand-off representations under strict numeric and categorical constraints. Results show that representation choice at the hand-off substantially influenced downstream reliability under a small decision model. Schema-constrained JSON extraction achieved the highest feasibility accuracy (0.96) and eliminated false-\texttt{NONE} outcomes in this setting, while narrative summarization, despite producing the smallest hand-off payload, substantially degraded feasibility (0.48) and increased false-\texttt{NONE} decisions. Embedding-based pruning matched the uncompressed control on feasibility (0.88) without an additional generative compression call. Selective preservation of original evidence provides a competitive baseline in this setting.

These findings show that compression should not be treated as a length reduction step alone. Constraint verification benefited from hand-off formats that retained explicit, auditable evidence, even when the resulting payload was not the smallest. 

\section{Future Areas of Research}
Future work will further evaluate mixed-model relays, larger inventories, and additional task families to test how these trade-offs scale with candidate set size and more open-ended goals.

\subsection{Mixed-model relays}
We plan to vary the Researcher and Booker model capacities independently to measure how hand-off formats interact with downstream capability and whether structured hand-offs remain advantageous when the Booker is stronger. 

\subsection{Long-context scaling and robustness}
Exploring how hand-off representations behave as the upstream trace grows (longer inventories, more constraints, and longer audits). This includes regimes where long-context models exhibit retrieval and attention failures in extended inputs. This would quantify whether structured hand-offs continue to outperform raw traces as context length increases, and whether compression can mitigate long-context degradation.

\subsection{Task generality}
Evaluating additional closed-world decision tasks with strict numeric and categorical constraints (e.g., scheduling and resource allocation) to test whether the observed feasibility-compression trade-offs persist across domains and constraint structures.

\bibliographystyle{IEEEtran}
\bibliography{references}

\end{document}